\documentclass{article}

\usepackage{arxiv}

\usepackage[utf8]{inputenc}
\usepackage[T1]{fontenc}
\usepackage{hyperref}
\usepackage{url}
\usepackage{booktabs}
\usepackage{amsfonts}
\usepackage{nicefrac}
\usepackage{microtype}
\usepackage{lipsum}
\usepackage{graphicx}
\usepackage{doi}
\usepackage{booktabs}
\usepackage{threeparttable}
\usepackage{makecell}
\usepackage{tcolorbox}
\usepackage[section]{placeins}
\usepackage{fontawesome5}
\tcbuselibrary{listings, raster, breakable}

\usepackage{pifont}
\newcommand{\cmark}{\ding{51}}%
\newcommand{\xmark}{\ding{55}}%

\usepackage{acro}
\NewDocumentCommand\acrodef{ m O{#1} m O{} }{%
  \DeclareAcronym{#1}{short={#2}, long={#3}, #4}%
}

\usepackage{tikz}
\usetikzlibrary{arrows.meta, positioning, fit, backgrounds, calc, shapes.geometric}

\usepackage{amsmath}
\usepackage{cleveref}

\newtcblisting{SmallInstructionbox}[1]{
    breakable,
    colback=black!4!white,
    colframe=black!75!white,
    coltitle=white,
    title={#1},
    arc=2mm,
    boxrule=0.5mm,
    left=3mm, right=3mm, top=3mm, bottom=3mm,
    listing only, 
    listing options={
        basicstyle=\small\rmfamily,
        breaklines=true,
        breakindent=0pt,
        columns=fullflexible,
        aboveskip=0pt,
        belowskip=0pt
    }
}

\newtcblisting{NonBreakableInstructionbox}[1]{
    colback=black!4!white,
    colframe=black!75!white,
    coltitle=white,
    title={#1},
    arc=2mm,
    boxrule=0.5mm,
    left=3mm, right=3mm, top=3mm, bottom=3mm,
    listing only, 
    listing options={
        basicstyle=\small\rmfamily,
        breaklines=true,
        breakindent=0pt,
        columns=fullflexible,
        aboveskip=0pt,
        belowskip=0pt
    }
}

\acrodef{PDDL}{Planning Domain Definition Language}
\acrodef{LM}{Language Models}
\acrodef{LLM}{Large Language Model}
\acrodef{NL}{Natural Language}
\acrodef{AI}{Artificial Intelligence}
\acrodef{FD}{Fast Downward}
\acrodef{TAMP}{Task and Motion Planning}
\acrodef{VRAM}{Video Random Access Memory}
\acrodef{GPU}{Graphics Processing Unit}
\acrodef{CRU}{Create Read Update}
\acrodef{COT}{Chain of Thought}

\title{PDDLCoder: Agentic PDDL Generation for LLM-Assisted Symbolic Planning}

\date{}

\author{Veit Laule \\
	Technical University of Berlin \\
    Berlin, Germany \\
	\texttt{v.laule@campus.tu-berlin.de} \\
	\And
	Jiangtao Shuai \\
	Technical University of Berlin \\
    Berlin, Germany \\
	\texttt{jiangtao.shuai@tu-berlin.de} \\
    \And
    Manfred Hauswirth \\
    Fraunhofer FOKUS \& Technical University of Berlin \\
    Berlin, Germany \\
    \texttt{manfred.hauswirth@tu-berlin.de}
    \And
    Sonja Schimmler \\
    Fraunhofer FOKUS \& Technical University of Berlin \\
    Berlin, Germany \\
    \texttt{sonja.schimmler@tu-berlin.de}
}

\renewcommand{\shorttitle}{PDDLCoder}

\hypersetup{
pdftitle={PDDLCoder: Agentic PDDL Generation for LLM-Assisted Symbolic Planning},
pdfsubject={cs.AI},
pdfauthor={Veit Laule, Jiangtao Shuai, Manfred Hauswirth, Sonja Schimmler},
pdfkeywords={AI Planning, Neuro-Symbolic Architecture, Agentic AI},
}

\begin{document}
\maketitle

\begin{abstract}
LLMs remain unreliable for long-horizon planning, often generating logically inconsistent or non-applicable plans.
Recent hybrid methods instead translate natural language into the Planning Domain Definition Language (PDDL), allowing symbolic planners to produce verifiable plans.
However, existing methods frequently rely on rigid generation pipelines, a partial PDDL definition, or human feedback.
Furthermore, their evaluation is hindered by the lack of standardized benchmarks with automated verification.
To address these limitations, we present PDDLCoder, an agentic framework for PDDL generation from natural language that iteratively generates, analyzes, and refines planning specifications. 
We further introduce NL-pddlgym, a benchmark dataset comprising 711 planning problems across 23 domains with executable gym environments for the automated verification of plan applicability. 
Experiments on the NL-pddlgym test set containing 106 problems across 4 held-out domains show that PDDLCoder generates applicable plans for 89.6\% of tested planning problems. 
This improves upon our adaptations of previous PDDL generation methods, which achieved up to 45.3\%, and outperforms direct LLM planning approaches, which reached up to 74.5\% on the same test set.
Our work demonstrates the effectiveness of agentic PDDL generation for planning and establishes a reproducible benchmark for future research on LLM-assisted symbolic planning.
\end{abstract}

\begin{center}
\faGithub~\href{https://github.com/vDawgg/PDDLCoder}{\nolinkurl{github.com/vDawgg/PDDLCoder}}
\end{center}


\section{Introduction}

Automated planning concerns the problem of finding a sequence of actions that transforms an initial state of an environment 
into one satisfying a desired goal condition \cite{fikes1971strips}. 
Unlike single-step decision problems, planning requires reasoning over long action sequences, where each action changes the state on which subsequent decisions depend. 
Therefore, reliable planning requires both an accurate representation of the environment and the ability to reason about the consequences of actions before execution.

With their broad world knowledge and ability to interpret \acf{NL}, 
\acp{LLM} have recently motivated new approaches for automated planning.
One intuitive paradigm, \textit{LLM-Planners}, uses \acp{LLM} directly to generate action sequences from \ac{NL} task descriptions~\cite{ahn2022can}.
Although such approaches demonstrate promising capabilities, they require the model to simultaneously infer the environment dynamics, reason over long horizons, and maintain action validity.
Recent studies show that \acp{LLM} often fail in this setting, producing plans containing actions that violate preconditions or do not reach a goal state \cite{huang2022language, valmeekam2023planning}.
Moreover, without an explicit representation of the planning problem, generated plans lack a reliable basis for validation before execution.

An alternative paradigm, LLM-Formalizers, shifts the role of LLMs from direct planning to formal model construction. 
These methods translate NL descriptions into formal planning representations, such as the \ac{PDDL} \cite{pddl}, which specify actions, preconditions, and effects. 
A symbolic planner such as \ac{FD} \cite{helmert2006fast} can then generate plans with formal guarantees relative to the generated model, provided that the model faithfully reflects the NL task. 
We view this formalization process as related to structured code generation, where LLM-based methods have demonstrated strong capabilities \cite{yang2024swe}.

Despite this potential, existing LLM-Formalizers remain limited in three aspects. 
First, many approaches rely on predefined generation pipelines with fixed refinement procedures \cite{oswald2024large, gestrin2024nl2plan}. 
Second, some methods simplify the task by assuming access to partial formal specifications, 
such as predefined domains \cite{llm+p}, or by incorporating human feedback during generation \cite{guan2023leveraging}. 
Third, evaluation often focuses on intermediate properties, such as syntactic validity \cite{hu-etal-2025-text2world}, 
rather than verifying whether generated specifications produce plans that successfully solve the original task.

To address these limitations, we introduce PDDLCoder, an agentic framework for \ac{PDDL} generation that iteratively creates, refines, and validates planning specifications through interaction with external tools. 
Unlike fixed pipelines, PDDLCoder autonomously determines refinement steps and operates zero-shot from \ac{NL} descriptions and action schemas without human intervention or access to predefined \ac{PDDL} domains, problems, preconditions, effects or predicates. 
We further introduce NL-pddlgym, a benchmark of 711 planning problems across 23 domains with executable environments for directly evaluating plan applicability. 
Our experiments show that PDDLCoder generates applicable plans for 89.6\% of the tested planning problems. 
This outperforms our adaptations of previous LLM-Formalizers, which achieved up to 45.3\%, and LLM-Planners, which reached up to 74.5\%.

\begin{figure*}[tbp]
    \centering
    \resizebox{\textwidth}{!}{%
    \begin{tikzpicture}[
        >=Stealth,
        font=\sffamily,
        input/.style={draw=black!70, rectangle, fill=gray!10, text width=3.2cm, align=center, minimum height=1.2cm, thick},
        agent/.style={draw=blue!60!black, rectangle, fill=blue!10, text width=2.8cm, align=center, minimum height=1.2cm, rounded corners, thick},
        document/.style={draw=black!70, rectangle, fill=gray!15, text width=2.8cm, align=center, minimum height=1cm, thick},
        mapnode/.style={draw=purple!60!black, rectangle, fill=purple!20, text width=3.5cm, align=center, minimum height=1cm, rounded corners, thick},
        gym/.style={draw=orange!60!black, rectangle, fill=orange!15, text width=3.5cm, align=center, minimum height=1.2cm, rounded corners, thick},
        database/.style={draw=black!70, cylinder, shape border rotate=90, aspect=0.25, fill=gray!20, text width=3cm, align=center, minimum height=1.3cm, thick},
        success/.style={draw=green!60!black, rectangle, fill=green!20, text width=3.5cm, align=center, minimum height=0.8cm, rounded corners, thick},
        failure/.style={draw=red!60!black, rectangle, fill=red!20, text width=3.5cm, align=center, minimum height=0.8cm, rounded corners, thick},
        toolcontainer/.style={draw=black!30, fill=cyan!10!gray!30, rounded corners, thick, minimum width=9.4cm, minimum height=4cm},
        tool/.style={draw=teal!60!black, rectangle, fill=teal!10, text width=2.6cm, align=center, minimum height=0.9cm, rounded corners, thick},
        pddlbox/.style={draw=black!30, fill=gray!5, rounded corners, inner sep=1pt, thick},
        gymbox/.style={draw=orange!50, fill=orange!5, rounded corners, inner sep=1pt, thick}
    ]
        
        \node[input] (prompt) at (0, 0) {\textbf{Natural Language \\ Descriptions} \\ ($D_{NL}, P_{NL}$)};
        
        \node[agent] (agent) at (4, 0) {\textbf{Agent} \\ (LLM)};
        \draw[->, thick] (prompt) -- (agent);
        
        \node[toolcontainer] (toolscontainer) at (11.5, 0) {};
        \node[above, font=\bfseries] at (toolscontainer.north) {\textbf{Tools}};
        
        \node[tool] (create) at (8.4, 1.2) {\textbf{Create PDDL File}};
        \node[tool] (edit) at (11.5, 1.2) {\textbf{Edit PDDL File}};
        \node[tool] (read) at (14.6, 1.2) {\textbf{Read PDDL File}};
        
        \node[tool] (syntax) at (9.9, -0.1) {\textbf{Get Syntax Errors}};
        \node[tool] (genplan) at (13.1, -0.1) {\textbf{Generate Plan}};
        
        \node[tool] (feedback) at (11.5, -1.3) {\textbf{Get Plan Feedback}};
        
        \draw[->, thick] ($(agent.east) + (0, 0.2)$) -- ($(toolscontainer.west) + (0, 0.2)$);
        \draw[<-, thick] ($(agent.east) + (0, -0.2)$) -- ($(toolscontainer.west) + (0, -0.2)$);
        
        \node[tool] (mapschema) at (4, -2) {\textbf{Map Plan Schema}};
        \node[document] (plan) at (4, -4) {\textbf{Plan}};
        
        \draw[->, thick] (agent) -- (mapschema);
        \draw[->, thick] (mapschema) -- (plan);
        
        \node[document] (pddlfiles) at (11.5, -4) {\textbf{PDDL Files}};

        \draw[->, thick] (pddlfiles.west) -| (6.3, -2) -- (mapschema.east);
        
        \draw[->, thick] ($(toolscontainer.south) + (-0.2, 0)$) -- ($(pddlfiles.north) + (-0.2, 0)$);
        \draw[<-, thick] ($(toolscontainer.south) + (0.2, 0)$) -- ($(pddlfiles.north) + (0.2, 0)$);
        
        \begin{scope}[on background layer]
            \coordinate (pddl_top) at ($(toolscontainer.north) + (0, 0.8cm)$);
            \coordinate (pddl_bot) at ($(plan.south) + (0, -0.4cm)$);
            \coordinate (pddl_right) at ($(toolscontainer.east) + (0.3cm, 0)$);
            \coordinate (pddl_left) at ($(prompt.west) + (-0.3cm, 0)$);
            
            \node[pddlbox, fit=(prompt) (plan) (pddlfiles) (toolscontainer) (pddl_top) (pddl_bot) (pddl_left) (pddl_right)] (pddlcoder) {};
            \node[above, font=\bfseries] at (pddlcoder.north) {\textbf{PDDLCoder}};
        \end{scope}
        
        
        \node[database] (dataset) at (19, 0.5) {\textbf{Dataset} \\ (23 Domains) \\ (711 Problems)};
        \node[gym] (gymenv) at (19, -3.3) {\textbf{Gym Environment} \\ (pddlgym)};
        \node[success] (applicable) at (23.8, -2.5) {\textbf{Applicable Plan} \\ (Success $\rightarrow$ Level 3)};
        \node[failure] (notapplicable) at (23.8, -4) {\textbf{Failure} \\ (Level 0-2)};
        
        \draw[->, thick] (dataset) -- node[right, font=\sffamily] {Simulates Problem} (gymenv);
        \draw[->, thick] (gymenv.east) -- (applicable.west);
        \draw[->, thick] (gymenv.east) -- (notapplicable.west);
        
        \begin{scope}[on background layer]
            \coordinate (gym_top) at ($(dataset.north) + (0, 0.4cm)$);
            \coordinate (gym_bot) at ($(notapplicable.south) + (0, -0.4cm)$);
            \coordinate (gym_left) at ($(gymenv.west) + (-0.3cm, 0)$);
            \coordinate (gym_right) at ($(notapplicable.east) + (0.3cm, 0)$);
            
            \node[gymbox, fit=(dataset) (gymenv) (applicable) (notapplicable) (gym_top) (gym_bot) (gym_left) (gym_right)] (nlpddlgym) {};
            \node[above, font=\bfseries] at (nlpddlgym.north) {\textbf{NL-pddlgym}};
        \end{scope}
        
        
        \draw[->, thick, dashed, draw=black!70] 
            (dataset.north) 
            -- (19, 3.8) 
            -- node[above, font=\bfseries] {Provides Problem Descriptions} (0, 3.8) 
            -- (prompt.north);
        
        \draw[->, line width=1.8pt, blue!60!black] 
            (plan.south) 
            -- (4, -5.5) 
            -- node[below, font=\bfseries] {Verify Plan} (19, -5.5) 
            -- (gymenv.south);

    \end{tikzpicture}%
    }%
    \caption{Conceptual overview of the main PDDLCoder pipeline and the interaction with NL-pddlgym. PDDLCoder receives a set of descriptions from NL-pddlgym for which the main agent then tries to generate an appropriate set of \ac{PDDL} files using the available tools. Once this is done, the plan can be mapped to its simulation-ready form and tested for applicability in the matching NL-pddlgym gym environment.}
    \label{fig:approaches}
\end{figure*}
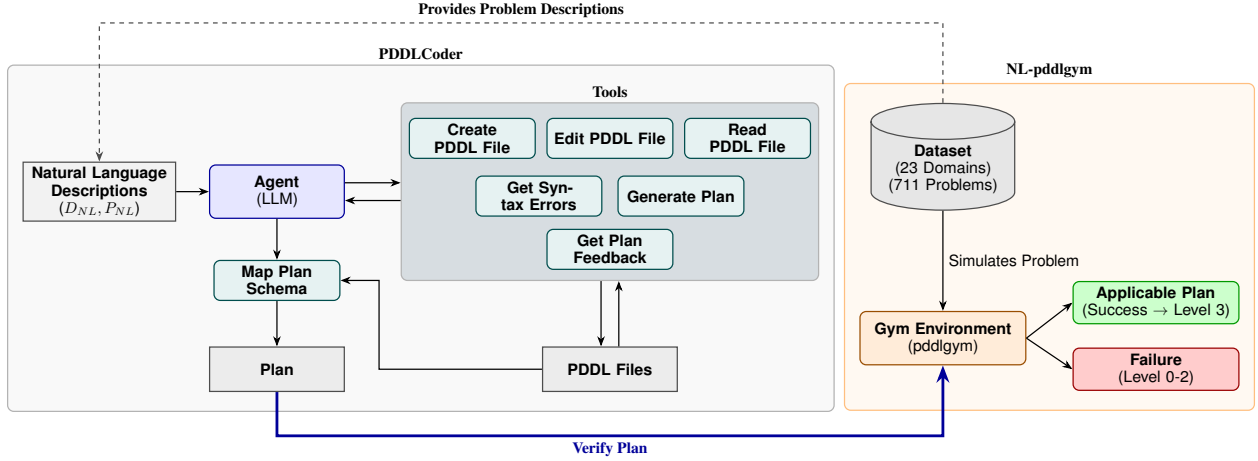

\section{Related Work}

\subsection{LLM-Planners}

With the advent of \acp{LLM}, researchers have explored the ability of these models to directly generate plans that solve a given problem.
Initial work explored the general reasoning capability of these models by applying \acp{LLM} to problems that typically require reasoning through techniques such as \ac{COT} prompting \cite{wei2022chain}.
Other work \cite{silver2022pddl} utilizes \ac{PDDL} as a means of unambiguously defining a given task, which is then provided as input to an \ac{LLM} to generate a plan.
This idea was further extended by using \acp{LLM} to generate intermediate \ac{PDDL} formalizations of the problem which are then used by the same model to generate a plan from \cite{zhou2024isr, pmlr-v235-kambhampati24a}.
Although some success in plan generation from \ac{PDDL} was reported, \acp{LLM} as \ac{PDDL}-Planners are still outperformed by dedicated planning systems specifically created for solving \ac{PDDL} problems.
In addition to these shortcomings, further work \cite{valmeekam2023planning}, which explored the long-term planning ability of off-the-shelf \acp{LLM} using \ac{NL} input, found that the models still generally struggle to generate applicable long-horizon plans.

\subsection{LLM-Formalizers}

In parallel to the exploration of LLM-Planners, researchers started generating formal representations, e.g. \ac{PDDL}, from \ac{NL} descriptions, which established planning methods then solve.
While LLM-Formalizers can use any number of intermediate formal representations of a given problem such as Python \cite{kagitha-etal-2026-unifying}, we focus on \ac{PDDL} as the formal representation in this work.
Early work \cite{llm+p} generated \ac{PDDL} problems from \ac{NL} descriptions but required a predefined \ac{PDDL} domain together with matching example pairs in the prompt.
Later work loosened these input requirements and additionally provided the model with feedback, such as syntactic checks whose results are incorporated back into the generated \ac{PDDL} \cite{guan2023leveraging}.

Subsequent work has improved complementary parts of the construction process.
To reduce context-window pressure, NL2PDDL \cite{oswald2024large} decomposes domain reconstruction into action-level generation steps.
VML\_PDDL \cite{yu2025generating} instead applies test-time scaling, selecting from multiple domain candidates before a fixed optimizer--learner refinement schedule, while its problem-synthesis experiments remain separate from domain synthesis.
NL2Plan \cite{gestrin2024nl2plan} broadens the target to end-to-end domain-and-problem generation from minimal text, but retains a staged extraction procedure followed by one repair pass.

\begin{table*}[htbp]
    \centering
    \begin{tabular}{llccccc}
        \toprule
        \textbf{Work} & \textbf{Role} & \textbf{ANL} & \textbf{D+P} & \textbf{AG} & \textbf{AR} & \textbf{VA} \\
        \midrule
        PlanBench \cite{NEURIPS2023_7a92bcde} & Benchmark & \xmark & -- & -- & -- & \cmark \\
        Text2World \cite{hu-etal-2025-text2world} & Benchmark & \xmark & \xmark & -- & -- & \xmark \\
        LLM+P \cite{llm+p} & Method & \xmark & \xmark & \cmark & \xmark & \xmark \\
        LLM-WorldModels \cite{guan2023leveraging} & Method & \cmark & \xmark & \xmark & \xmark & \xmark \\
        NL2PDDL \cite{oswald2024large} & Method & \xmark & \xmark & \cmark & \xmark & \xmark \\
        NL2Plan \cite{gestrin2024nl2plan} & Method & \cmark & \cmark & \cmark & \xmark & \xmark \\
        VML\_PDDL \cite{yu2025generating} & Method & \cmark & \xmark & \cmark & \xmark & \xmark \\
        \midrule
        NL-pddlgym (ours) & Benchmark & \cmark & \cmark & -- & -- & \cmark \\
        PDDLCoder (ours) & Method & \cmark & \cmark & \cmark & \cmark & \cmark \\
        \bottomrule
    \end{tabular}%
    \caption{Method and benchmark boundaries. Abstract \ac{NL} (ANL) means that task descriptions are abstract and full predicate descriptions are not supplied; Domain and Problem (D+P) means joint domain-and-problem generation; Autonomous Generation (AG) means that neither human feedback nor oracle environments are needed during generation; Adaptive Refinement (AR) means that the method is not limited to rigid refinement schedules; Verified Applicability (VA) means that an output plan is executed against task dynamics, rather than assessed only by syntax, structural similarity, or agreement with a reference. Dashes denote a non-applicable method property.}
    \label{tab:related_comparison}
\end{table*}

These developments improve decomposition, candidate selection, and task coverage, but Table \ref{tab:related_comparison} highlights two remaining distinctions.
First, their refinement schedules are specified in advance rather than selected in response to the current failure.
PDDLCoder addresses this workflow limitation by treating syntax errors, unsolvable \ac{PDDL} models, and solvable models whose plans conflict with the task semantics as different states requiring different tools.
To achieve this, the agent chooses its next action from the evolving workspace instead of applying the same repair sequence to every instance.
Second, formal validity or symbolic solvability alone does not establish that a generated plan solves the task expressed in \ac{NL}.
We address this separate evaluation gap with NL-pddlgym, which executes the final mapped plan against held-out task dynamics, as described next.

\subsection{Benchmarks for Language-to-PDDL Formalization}

Existing benchmarks emphasize different stages of the formalization-to-execution pipeline.
Planetarium \cite{zuo2025planetarium} provides a large collection of text-to-\ac{PDDL} problem pairs and a semantic-equivalence test, but assumes fixed domains and is limited to three classical domains.
Text2World \cite{hu-etal-2025-text2world} broadens domain coverage and evaluates generated domain models with syntactic validity, solvability, structural, and component-level metrics, but does not verify plan-applicability, supplies predicate descriptions and does not require a matching problem file.

PlanBench \cite{NEURIPS2023_7a92bcde} shifts attention to the downstream stage: it uses planner- and validator-based checking to evaluate direct planning and reasoning about change.
Its few-shot prompts specify the lifted actions, preconditions, and effects, leaving domain construction outside the task.
Our NL-pddlgym connects formalization to this downstream evaluation.
It provides a high-level task description, action names, objects, initial states, and goals.
Separately, the system receives the minimal target action schema S needed to map generated plans to the interface of the target environment. 
The model must infer the withheld predicates, preconditions, effects, and any additional parameters needed by the generated \ac{PDDL}.
This design follows the argument of \cite{huang2025planning} that expert-crafted formalization hints both limit scalability and bias the interpretation of inherently ambiguous \ac{NL} descriptions.
The resulting plan is then mapped and executed in a corresponding gym environment.
NL-pddlgym thus combines low-specification joint domain and problem formalization with plan applicability evaluation while remaining complementary to Text2World's domain-model analysis and PlanBench's broader planning curriculum.

\section{Approach}

We study joint domain-and-problem formalization to generate plans for fully observable and deterministic classical planning problems \cite{ghallab2004automated}.
Given abstract \ac{NL} descriptions of a planning domain and a problem, denoted by $D_{NL}$ and $P_{NL}$, respectively, PDDLCoder uses an \ac{LLM} and symbolic feedback tools to construct a \ac{PDDL} domain $D$ and problem $P$ (see Section \ref{sec:pddlcoder}), from which a classical planner derives a plan $\pi$.
The generation process receives neither human corrections nor feedback from the held-out gym environment.

For the final applicability evaluation we do not consider the original $\pi$ resulting from a given $D$ and $P$, but a mapped version $\pi_m$, which matches the action schema $S$ and object set $O$ expected in the gym environment.
$S$ contains actions with the set of parameters that are deemed absolutely necessary to deterministically carry out an action in each domain as in the original pddlgym implementation \cite{silver2020pddlgym}.
In $S$, each schema is specified using only its action and parameter names, restricting the amount of domain-specific information provided.
For example, a 'move' operator in \ac{PDDL} is often parameterized with a \textit{from} location and a \textit{to} location.
While it makes sense to supply the original \textit{from} location in \ac{PDDL}, as this easily allows for removing the old location from the current set of facts, an agent moving to a new location does not need this information, as it only needs to know which new location to move to \cite{silver2020pddlgym}.
This simplified 'move' action would then be given as 'move(?to)' in $S$.
By working with this minimal set of parameters for each action, we allow our approach to freely choose the parameters that are used in the generated \ac{PDDL} while assuring that we can later test the applicability of $\pi_m$ in the gym environments without encountering issues due to the arity of the generated actions.
To generate an appropriate $\pi_m$ from a given $\pi$ we use an \ac{LLM} that we call \textit{MapAgent} which receives, $S$, $O$ and the generated $D$ and $\pi$ to generate the matching $\pi_m$.
We leave probabilistic domains and other extensions to \ac{PDDL} for future work.

\subsection{NL-pddlgym}

We introduce NL-pddlgym, a benchmark for generating both \ac{PDDL} domain and problem files from \ac{NL} descriptions and evaluating the resulting plans through execution.
It comprises 711 problems across 23 domains, with a median of 30 problems per domain and between 3 and 310 objects per problem instance, and pairs each $(D_{NL}, P_{NL})$ input with an executable environment.
Our experiments reserve four complete domains as a held-out test set and the remaining domains form training and validation splits for future method development, with no test domain appearing in either split.
The executable environments are built with pddlgym, which provides gym interfaces for \ac{PDDL} domains and problems \cite{silver2020pddlgym}.
For each generated $\pi_m$ the evaluator applies the plan's actions sequentially, and records success only if all actions could be applied, meaning their preconditions were satisfied, and the terminal state satisfies the task goal.
NL-pddlgym therefore supplies a task-level applicability oracle rather than relying only on syntax checks or symbolic solvability.
Furthermore, we do not make any claims regarding the semantic equivalence of the generated and ground-truth PDDL.
Due to the abstract definition of the planning problem in NL and the explicit omission of predicates from this description, NL-pddlgym allows for multiple valid PDDL encodings.
We therefore limit the benchmark to testing the applicability of the plans resulting from the \ac{PDDL} files for a given problem.

The underlying \ac{PDDL} domains and problems were drawn from the original pddlgym implementation and the classical-domains repository\footnote{classical-domains GitHub repository \url{https://github.com/AI-Planning/classical-domains}}, complemented by three domains and their problems created specifically for this work.
We restrict all domains to the requirements \textit{typing}, \textit{strips}, \textit{disjunctive-preconditions}, \textit{conditional-effects}, \textit{negative-preconditions} and \textit{equality}.
To construct the language side of the benchmark, we manually authored all 23 domain descriptions and two to three seed problem descriptions per domain, then used DeepSeek v4 Flash \cite{deepseek_v4} to generate the descriptions for the remaining problems, instructing the model to follow the seed patterns while preserving the objects, initial state, and goal of the source \ac{PDDL} problem.
Each domain description summarizes the task setting and lists the available action names, while each problem description specifies the objects, initial state, and goal.
To preserve the abstract formalization setting, $D_{NL}$ and $P_{NL}$ omit predicate signatures, action parameters, preconditions, and effects.
In addition to this, the dataset also contains an object list $O$ and action schema list $S$ needed to map generated plans $\pi$ to their simulation-ready form $\pi_m$.
An example of $D_{NL}$ and $P_{NL}$ for the Ring and Peg domain from the test set can be seen in Figure \ref{fig:example_prompts}.

\begin{figure}
\begin{tcbraster}[raster columns=2, raster equal height, raster column skip=1em]
\begin{NonBreakableInstructionbox}{Ring and Peg Domain Description}
## Domain description

A robot arm is tasked to sort a set of colored rings on a set of pegs of the same colors. The arm can be used to pick up a ring from a peg, move a ring to another peg and place the ring on the other peg. Rings are already placed on pegs and can only be placed on pegs. To pick up a ring at a peg, the robot first has to move to the pegs position.

The actions available to the robot are:
- move
    - Moves the robot to a specified position
- pick
    - Closes the robot arms gripper at the current position
- place
    - Opens the robot arms gripper at the current position
\end{NonBreakableInstructionbox}%
\begin{NonBreakableInstructionbox}{Exemplary Ring and Peg Problem Description}
## Problem description

There are 5 colored pegs with the following colors: red, green, blue, pink, yellow. The pegs are all named in the format <color>_peg. Additionally there are 3 colored rings: red, green and blue. Similarly to the pegs, the rings are named <color>_ring.

The robot arm starts out in a default position, while the rings start in the following positions:
- red_ring - pink_peg
- green_ring - yellow_peg
- blue_ring - red_peg

The goal is to transfer all rings to the pegs of their color.
\end{NonBreakableInstructionbox}
\end{tcbraster}
\caption{Exemplary $D_{NL}$ and $P_{NL}$ of the Ring and Peg domain from the NL-pddlgym test set}
\label{fig:example_prompts}
\end{figure}

\subsection{PDDLCoder}
\label{sec:pddlcoder}

Figure \ref{fig:approaches} depicts the control flow and system architecture of PDDLCoder. 
Orchestrated via the DSPy \cite{khattab2024dspy} implementation of ReAct \cite{yao2022react}, the agent receives the natural language descriptions ($D_{NL}$, $P_{NL}$) and general \ac{PDDL} guidelines to iteratively construct the domain $D$ and problem $P$. 
The tools the agent has access to can be broadly divided into two categories: the first allows for the creation, reading, and editing of \ac{PDDL} files, while the second allows the agent to gather feedback on the current state of the generated files.
The agent continuously interacts with these tools until it explicitly signals completion or reaches a hard limit of 50 iterations, which prevents context window exhaustion. 
Following loop termination, the system validates the syntax of $D$ and $P$ and attempts to generate a plan $\pi$. 
If successful, $\pi$ is mapped to $\pi_m$ by the MapAgent and saved.
Otherwise, a syntax error or planning failure is recorded.

\subsection{\ac*{CRU} Tools}
\label{sec:cru_tools}

The \ac{CRU} tools we use in the pipeline are similar to part of the code-editing interfaces outlined in \cite{yang2024swe}, with the main difference being that the operations in our approach are specifically limited to working with \ac{PDDL} files. 
While the creation and reading tools directly manipulate files, the edit tool first applies modifications in memory and validates them using the VAL parser \cite{howey2004val}. 
If syntax errors are detected, the edit is rejected and reported back.
Otherwise, the updated file contents are returned.

\subsection{Feedback Tools}
\label{sec:feedback_tools}

The tools in this category provide feedback on the quality of the generated \ac{PDDL} using external programs, such as VAL \cite{howey2004val} or \ac{FD} \cite{helmert2006fast}, allowing the agent to check syntax, generate plans, or assess plan applicability.

\subsubsection{Syntax Check Tools}
\label{sec:syntax_check_tools}

The syntax tools leverage VAL to report errors with line numbers and file contexts after the initial creation of $D$ or $P$. 
To complement VAL, we also provide a tool running the translation phase of \ac{FD} \cite{helmert2006fast}. 
While VAL offers detailed error diagnostics and line numbers, \ac{FD}'s translation step acts as a stricter syntax check required before plan generation.

\subsubsection{Plan Generation Tools}
\label{sec:plan_generation_tools}

Once a set of \ac{PDDL} files has been created, the agent can call a tool to verify if $D$ and $P$ can be used to generate a plan, returning either $\pi$ or a planning-failure report.
For the plan generation, we use the \textit{lama-first} configuration of \ac{FD}, which, while not guaranteed to produce optimal plans, quickly returns the first plan it finds enabling fast feedback loops.
To support parallel execution, we cap the memory of each \ac{FD} run at 4 GB and additionally limit execution time to 1 minute.
When planning fails due to unsolvability, we parse a curated subset of \ac{FD}'s verbose output to extract what we consider to be the most actionable data.
This includes the number of reachable grounded \ac{PDDL} atoms, grounded actions, actions removed due to impossible preconditions and effects, and information on the satisfiability of the goal state.
In cases where planning fails due to time- or memory-limits, we simply return a message stating this information.

\subsubsection{Plan Feedback Tool}
\label{sec:plan_feedback_tools}

Once $\pi$ is generated, the agent can invoke a feedback tool to evaluate its task applicability, inspired by iterative refinement methods \cite{madaan2023self}. 
Rather than evaluating the raw plan $\pi$, this tool prompts a dedicated \textit{PlanAgent} using only the natural language descriptions ($D_{NL}$, $P_{NL}$), the target action schema $S$, and the mapped plan $\pi_m$. 
We intentionally withhold the generated $D$ and $P$ to prevent the PlanAgent from over-indexing on the syntactic design of the formal representations, ensuring its feedback remains focused on the semantic feasibility of the action sequence. 
The PlanAgent parses $\pi_m$ to identify violations of explicit or implicit constraints derived from $D_{NL}$. 
For example, in the \textit{Ring and Peg} domain, where a set of rings needs to be transferred between a set of pegs, \acp{LLM} frequently generate plans that try to move a ring while it is stacked underneath another ring on the same peg. 
The PlanAgent can identify this implicit physical violation in the mapped plan and provide actionable guidance to the main agent to correct the underlying \ac{PDDL}.

\section{Evaluation}

We evaluate PDDLCoder on the test set of the NL-pddlgym dataset.
This subset of the dataset consists of 4 domains not included in the remaining dataset splits, which are intended for optimization approaches.
The test set includes a domain (Ring and Peg) that we specifically created for this dataset, which to our knowledge has not been publicly available prior to these experiments.
The remaining domains are the established classical planning domains Hanoi, Elevator and Satellite.
The total number of problems in this test set amounts to 106 problem variations with varying numbers of objects, resulting in plans ranging between 2 and 557 steps.

To account for the influence of different \acp{LLM} and the impact of the different tools we designed for PDDLCoder, we evaluate the approach across a set of open-weight models and four ablations.
For these ablations we either remove all feedback tools, only the plan feedback tool, or the plan mapping step for the feedback tool.
This set of ablations is complemented by a rigid version of PDDLCoder to analyze the impact of the methods execution-style.
In addition to the ablations of our approach, we also include comparisons with adaptations of previous LLM-Formalizer and LLM-Planner methods.
The set of LLM-Formalizer methods we consider consists of NL2Plan \cite{gestrin2024nl2plan} and an adapted version of VML\_PDDL \cite{yu2025generating}, where we use LLM-generated scores for the BoN selection instead of selecting via log-probs as in the original paper due to restrictions of the DeepSeek API we used for the experiments.
For comparison with LLM-Planner methods, we evaluate the self-validator variant of ISR-LLM, which uses few-shot prompts from the Blocksworld domain \cite{zhou2024isr}, and a \ac{COT} \cite{wei2022chain} prompting approach on the same test set.
All of the approaches we compare with PDDLCoder were executed in their best-performing configurations including original prompts, candidate counts, iteration limits and stopping criteria as specified in the original papers.
To ensure a fair comparison with these approaches, we employ the same mapping approach for the generated plans as used in PDDLCoder.
Each model, problem and approach configuration was evaluated once.
The reported results therefore do not quantify sampling variance between different runs.

We separately verified the quality of the MapAgent approach by running the MapAgent on the plans generated from the ground-truth \ac{PDDL} files from the NL-pddlgym test set.
After applying each of the generated plans to the appropriate gym environments, we found that the mapped plans yielded a $98.1\%$ success rate in solving the problem given a correct initial unmapped plan. 

\begin{table}[htbp]
    \centering
    \begin{tabular}{ll}
        \toprule
        \textbf{Level} & \textbf{Description} \\
        \midrule
        0 & No syntactically valid \ac{PDDL}. \\
        1 & Syntactically valid \ac{PDDL} but no plan found. \\
        2 & Non-applicable plan. \\
        3 & Plan solves the given problem. \\
        \bottomrule
    \end{tabular}
    \caption{Levels for \ac{PDDL} generation.}
    \label{tab:pddl_levels}
\end{table}

To evaluate this varied set of approaches we use a metric we created for the \ac{PDDL} generation task. 
This metric is designed to measure the capability of a \ac{PDDL} generation approach along a set of four distinct levels, denoting different stages of increasing difficulty that need to be passed during \ac{PDDL} generation, which are described in more detail in Table \ref{tab:pddl_levels}.
Generally, level 0 indicates inability to generate syntactically valid \ac{PDDL}, meaning the generated PDDL was rejected by the VAL parser or \ac{FD} translation layer, and level 3 indicates the ability to generate \ac{PDDL} that produces applicable plans solving a given problem.
Notably, a plan is applicable iff each action is applicable in the given sequence and the execution of the terminal action satisfies the goal.
In addition to the LLM-Formalizer approaches we evaluate here, we also evaluate the LLM-Planner approaches using this metric.
However, here the metric only accounts for the generation of non-applicable or applicable plans (levels 2 and 3) as no \ac{PDDL} is generated using these methods.

\subsection{Results Across Models}

We initially present the results of our approach when applied to a variety of different open-weight models, to verify whether PDDLCoder scales to different \acp{LLM}.

\begin{table}[htbp]
    \centering
    \makebox[\textwidth][c]{
    \begin{tabular}{l|cccc|cccc|cccc}
        \toprule        
        & \multicolumn{4}{c|}{\textbf{Outcome Counts}} & \multicolumn{4}{c|}{\textbf{Success Rates (\%)}} & \multicolumn{4}{c}{\textbf{Token Statistics}} \\
        \textbf{Model} & \textbf{L0} & \textbf{L1} & \textbf{L2} & \textbf{L3} & \textbf{El} & \textbf{Ha} & \textbf{RP} & \textbf{Sa} & \textbf{N-In} & \textbf{S-In} & \textbf{N-Out} & \textbf{S-Out} \\
        \midrule
	    DeepSeek v4 Flash & 0 & 2 & 9 & \textbf{95} & \textbf{90} & 75 & \textbf{93.3} & \textbf{94.4} & 172978.8 & 234922.6 & 32809.4 & 32990.4 \\
        Gemma 4 31b & 8 & 2 & 31 & 65 & 20 & \textbf{85} & 63.3 & 69.4 & 411577.6 & 404900.1 & 15453.8 & 12257.4 \\
        gpt-oss-120b & 26 & 3 & 21 & 56 & 80 & 55 & 46.7 & 41.7 & 307318.1 & 424007.3 & 27589.4 & 29793.0 \\
        Qwen 3.6 35b A3b & 45 & 1 & 49 & 11 & 30 & 5 & 6.7 & 5.6 & \textbf{129916.1} & 251746.6 & 37574.6 & 40290.6 \\
        GLM 4.7 Flash  & 66 & 17 & 22 & 1 & 5 & 0 & 0 & 0 & 762909.5 & 721951.2 & 35162.3 & 28381.5 \\
        Llama 4 Scout 17b 16e & 99 & 3 & 4 & 0 & 0 & 0 & 0 & 0 & 521045.1 & \textbf{154822.6} & \textbf{13961.4} & \textbf{6312.5} \\
        \bottomrule
    \end{tabular}}
	\caption{Performance of PDDLCoder across different open-weight \acp{LLM} on the NL-pddlgym test set. Also showing the success rates (i.e. reached level 3) for the Elevator (El), Hanoi (Ha), Ring and Peg (RP) and Satellite (Sa) domains from the test set. Additionally includes the average number of input (N-In) and output (N-Out) tokens together with standard deviation (S-In and S-Out).}
    \label{tab:model_comparison}
\end{table}

In the case of DeepSeek v4 Flash, the model was used through OpenRouter due to the possibility of executing multiple experiments in parallel, while the other models were hosted locally using vLLM \cite{kwon2023efficient}.
Looking at the results in Table \ref{tab:model_comparison} we can see that, although a portion of the models performs well, the effectiveness of PDDLCoder is dependent on the ability of the \acp{LLM} to generate usable plan feedback and syntactically valid \ac{PDDL}.
Amid runs that produce syntactically valid PDDL, relatively few terminate at level 1. 
This suggests that planner solvability is less frequently a bottleneck than producing syntactically valid files or ensuring that the resulting plan is applicable.

Among the tested models, DeepSeek v4 Flash \cite{deepseek_v4} performs the best, generating \ac{PDDL} resulting in applicable plans for $89.6\%$ of the problems, followed by the smaller Gemma 4 31b \cite{gemma_4} and gpt-oss-120b \cite{agarwal2025gpt}.
The Gemma model performs surprisingly well for its size when compared to the 258 billion parameter DeepSeek model, and is able to generate syntactically valid \ac{PDDL} in $92.5\%$ of the cases.
However, the model frequently fails to point out issues with generated plans, leading to $29.2\%$ not being applicable versus $8.5\%$ of non-applicable plans among the tested problems for the DeepSeek model.
These results indicate that the quality of plan feedback is strongly model-dependent, although we cannot clearly state whether this variation is caused by model scale, training, or instruction-following ability.
More than this, some models like Qwen 3.6 35b A3b \cite{qwen36_35b_a3b} and gpt-oss-120b frequently fail to adhere to the expected response format, resulting in no \ac{PDDL} being generated for a given problem.
In other instances, the \acp{LLM} fail to generate solvable $D$ and $P$ for the Hanoi domain, as they are unable to adhere to the requirement stated in $D_{NL}$ that the destination of the move action must be a peg.
Here, the \acp{LLM} frequently start out with generating \ac{PDDL} that does not fit this constraint and are unable to generate \ac{PDDL} that is correctly aligned before the budget is exhausted.
Most notably, other models like GLM 4.7 Flash \cite{5team2025glm45agenticreasoningcoding} and Llama 4 Scout 17b 16e \cite{llama-4} are largely unable to generate syntactically valid \ac{PDDL} indicated by the models only generating \ac{PDDL} without syntax issues in $37.7\%$ or even $6.6\%$ of the cases respectively.
Specifically, the models fail to apply any fixes to the syntactically invalid \ac{PDDL} generated by them and frequently try to apply the same invalid edit until the iteration budget is exhausted.
Seeing this, a major bottleneck for PDDLCoder is the ability of an \ac{LLM} to generate syntactically valid \ac{PDDL} as this forms the basis for all later stages in the \ac{PDDL} generation process.

\subsection{Comparison with Previous Approaches}

Table \ref{tab:approach_comparison} compares PDDLCoder with adaptations of previous LLM-Formalizers and LLM-Planners.
PDDLCoder achieves a level-3 success rate of $89.6\%$, compared with $45.3\%$ for NL2Plan, $34.9\%$ for VML\_PDDL, $74.5\%$ for COT, and $69.8\%$ for ISR-LLM.
PDDLCoder achieves this performance while requiring a lower number of output tokens on average than other LLM-Formalizers, although it consumes substantially more input tokens due to long tool outputs with full \ac{PDDL} files or plans.

\begin{table*}[htbp]
    \centering
    \begin{tabular}{l|cccc|cccc|cccc}
        \toprule        
        & \multicolumn{4}{c|}{\textbf{Outcome Counts}} & \multicolumn{4}{c|}{\textbf{Success Rates (\%)}} & \multicolumn{4}{c}{\textbf{Token Statistics}} \\
        \textbf{Approach} & \textbf{L0} & \textbf{L1} & \textbf{L2} & \textbf{L3} & \textbf{El} & \textbf{Ha} & \textbf{RP} & \textbf{Sa} & \textbf{N-In} & \textbf{S-In} & \textbf{N-Out} & \textbf{S-Out} \\
        \midrule
        PDDLCoder & 0 & 2 & 9 & \textbf{95} & \textbf{90} & \textbf{75} & 93.3 & \textbf{94.4} & 172978.8 & 234922.6 & 32809.4 & 32990.4 \\
        \midrule
        VML\_PDDL & 52 & 4 & 13 & 37 & 75 & 10 & 33.3 & 27.8 & 83338.5 & 63977.7 & 123071.6 & 26272.6 \\
        NL2Plan & 34 & 5 & 19 & 48 & 70 & 40 & 33.3 & 44.4 & 149727.6 & 70625.3 & 44023.2 & 26275.4 \\
        \midrule
        ISR-LLM & - & - & 32 & 74 & 80 & 40 & 83.3 & 69.4 & 13467.3 & 10841.8 & 22758.5 & 19680.3 \\
        COT & - & - & 27 & 79 & \textbf{90} & 60 & \textbf{96.7} & 55.6 & \textbf{2473.7} & \textbf{2603.0} & \textbf{7410.1} & \textbf{8658.1} \\
        \bottomrule
    \end{tabular}
    \caption{Performance of PDDLCoder compared with adapted versions of previous LLM-Formalizers (VML\_PDDL and NL2Plan) and LLM-Planners (ISR-LLM and COT). All approaches were tested on the NL-pddlgym test set with DeepSeek v4 Flash.}
    \label{tab:approach_comparison}
\end{table*}

The improvements over VML\_PDDL \cite{yu2025generating} might be due to the lack of feedback mechanisms, which are present in PDDLCoder.
Although the model is also able to iteratively refine the generated \ac{PDDL} it cannot externally verify that the generated \ac{PDDL} is actually syntactically correct or solvable, leading to $49\%$ of the generated problems not passing syntax checks.

In contrast to this, NL2Plan \cite{gestrin2024nl2plan} offers even more feedback mechanisms to the model than PDDLCoder but might do so in a too rigid and narrow manner.
Specifically, the pipeline fails to generate syntactically valid \ac{PDDL} for $34$ of the planning problems from the test set, which most frequently stems from actions using non-existent predicates or ill-defined action names.
This is partially due to syntactically incorrect action names which cannot be changed after their initial generation.
While this approach also incorporates a form of plan feedback during generation, the feedback is only ever generated once.
This assumes that the initial feedback is perfect and that the issue can be resolved with this feedback alone, which is often not the case, as suggested by the large number of non-applicable plans.
In contrast to this, PDDLCoder allows for multiple calls of the plan feedback (within the total budget), allowing the model to iteratively verify that its changes resolved issues that were found by the PlanAgent.

\begin{figure}[htbp]
    \centering
    \includegraphics[width=.5\linewidth]{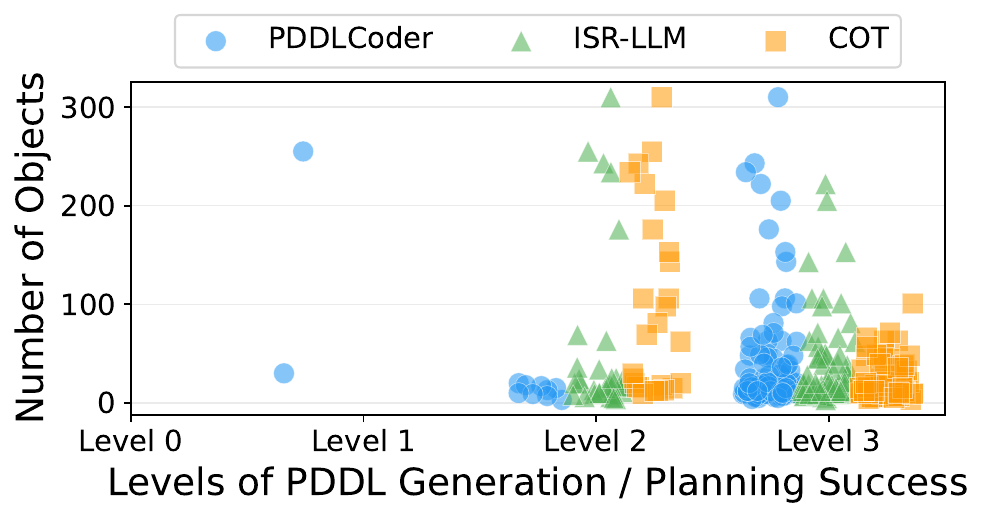}
    \caption{Relationship between object count and performance compared across PDDLCoder, COT, and ISR-LLM.}
    \label{fig:object_count_and_performance}
\end{figure}

Surprisingly, the LLM-Planner approaches we tested outperform all tested LLM-Formalizer approaches except PDDLCoder when used with DeepSeek v4 Flash on the NL-pddlgym test set, although this ranking does not always transfer to other models in our testing.
Even though ISR-LLM \cite{zhou2024isr} uses a more sophisticated pipeline design when compared to \ac{COT} \cite{wei2022chain} both LLM-Planner approaches perform comparably.
Notwithstanding its overall good performance, \ac{COT} still struggles with generating applicable plans for complex planning problems as already noted in previous work that looked into the planning ability of \acp{LLM} \cite{shojaee2025the, valmeekam2023planning}.
This becomes apparent when looking at Figure \ref{fig:object_count_and_performance}, where the number of objects, which we use as a proxy to the problem's complexity similarly to \cite{shojaee2025the}, is compared with the performance of the approaches.
Here, COT fails to generate applicable plans that involve more than 100 objects, due to missing required actions or the refusal to generate plans at all in sufficiently complicated problem instances.
However, a similar cut-off cannot be found for ISR-LLM, although this method still frequently fails to adhere to the required output format for the generated plan.
On the same test set, PDDLCoder is able to generate applicable plans for the tested problem instance with the highest object count.
Nevertheless, unlike PDDLCoder, the direct LLM-Planner approaches do not produce a reusable formal model from which a symbolic planner can deterministically regenerate plans.

\begin{figure}[htbp]
    \centering
    \includegraphics[width=\linewidth]{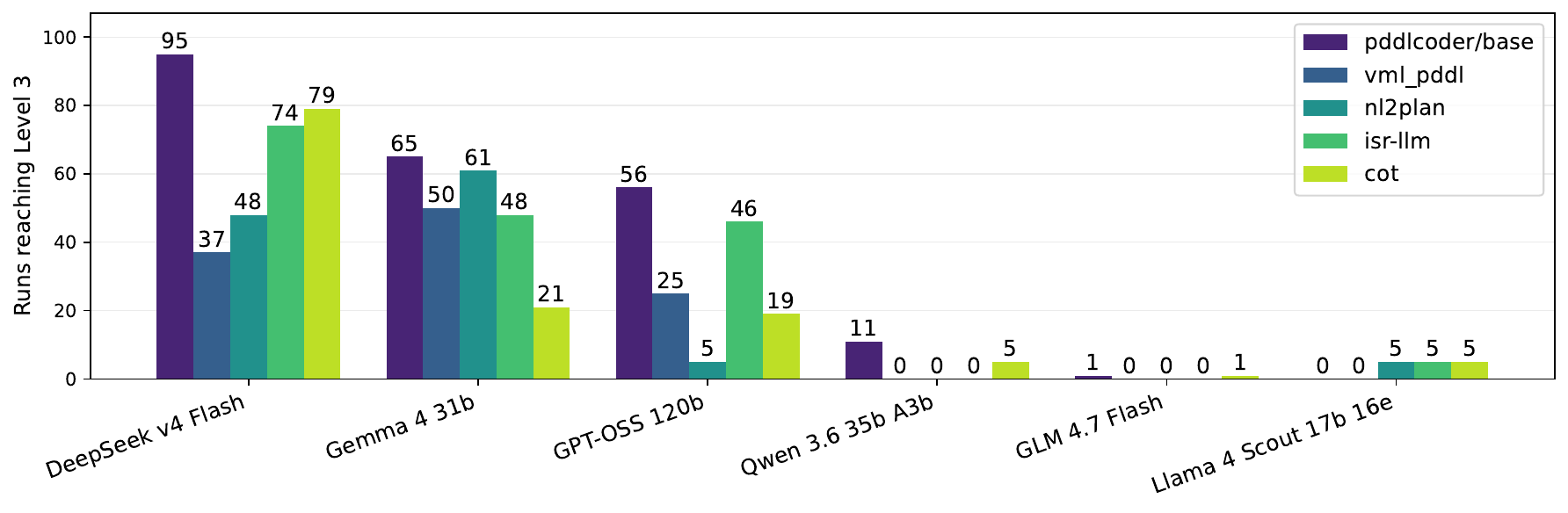}
    \caption{Performance (number of runs reaching level 3) across open-weight models and adaptations of previous LLM-Planner and LLM-Formalizer approaches. ISR-LLM was not evaluated with Qwen 3.6 35b A3b because the model does not support the intermediate system prompts required by the method. The corresponding zero-height bar denotes missing results rather than zero successful instances.}
    \label{fig:model_and_approach_perf}
\end{figure}

Finally, when looking at the performance of each of the distinct approaches across the open-weight models in Figure \ref{fig:model_and_approach_perf}, we can see that PDDLCoder continues to be the strongest approach, with the exception of the results of GLM 4.7 and Llama 4 Scout 17b 16e.
Performance varies substantially across models, and the relative ranking of the previous approaches is not consistent, highlighting the importance of model selection. 
Nevertheless, PDDLCoder achieves the highest or tied-highest level-3 performance for all models except Llama 4 Scout 17b 16e.

\subsection{Ablation Results}

As can be seen in Table \ref{tab:ablation_comparison} the tools we designed as part of PDDLCoder all positively influence the syntactic validity and solvabilty of the generated \ac{PDDL} and the applicability of generated plans.
Specifically, removing all feedback tools except for the \ac{CRU} tools from the pipeline results in PDDLCoder generating syntactically invalid \ac{PDDL} for $11\%$ of the tested problems, suggesting that the external feedback contributes to syntactic reliability.
Furthermore, we can see that the addition of the PlanAgent increases the number of applicable plans generated for the test set.
This is due to the approach often generating plans that do not respect logical constraints, such as stacking constraints that are not correctly inferred for the Ring and Peg domain described earlier, which can be pointed out by the PlanAgent.
However, this increase in the number of applicable plans is also accompanied by a $2.74x$ increase in the number of consumed output tokens, which is the largest among the tested agentic PDDLCoder ablations.
Additionally, we can see that while the usage of $\pi_m$ during plan feedback provides a noticeable improvement, it does not constitute a similar impact as the PlanAgent which results in a $20.8$ pp. decrease in level 3 outcomes when removed from the full pipeline, while the additional switch from $\pi_m$ to $\pi$ for plan feedback only results in a further $7.5$ pp. decrease.

\begin{table}[htbp]
    \centering
    \makebox[\textwidth][c]{
    \begin{tabular}{l|cccc|cccc|cccc}
        \toprule        
        & \multicolumn{4}{c|}{\textbf{Outcome Counts}} & \multicolumn{4}{c|}{\textbf{Success Rates (\%)}} & \multicolumn{4}{c}{\textbf{Token Statistics}} \\
        \textbf{Ablation} & \textbf{L0} & \textbf{L1} & \textbf{L2} & \textbf{L3} & \textbf{El} & \textbf{Ha} & \textbf{RP} & \textbf{Sa} & \textbf{N-In} & \textbf{S-In} & \textbf{N-Out} & \textbf{S-Out} \\
        \midrule
        PDDLCoder & 0 & 2 & 9 & \textbf{95} & 90 & \textbf{75} & \textbf{93.3} & \textbf{94.4} & 172978.8 & 234922.6 & 32809.4 & 32990.4 \\
        Rigid Execution & 19 & 2 & 25 & 60 & 50 & 60 & 46.7 & 66.7 & \textbf{39319.1} & \textbf{31494.2} & 57247.8 & 43709.1 \\
        Without MapAgent & 7 & 1 & 11 & 87 & 90 & 70 & 90 & 77.8 & 320124.2 & 355922.2 & 41477.0 & 31389.1 \\
        Without PlanAgent & 3 & 2 & 28 & 73 & 90 & 85 & 40 & 72.2 & 74907.8 & 65274.9 & \textbf{15127.5} & \textbf{11861.2} \\
        CRU Only & 12 & 7 & 19 & 68 & \textbf{95} & 60 & 36.7 & 72.2 & 48243.1 & 41198.7 & 18176.0 & 11773.6 \\
        \bottomrule
    \end{tabular}}
    \caption{Performance of PDDLCoder compared with a rigid version of the pipeline and ablations where the plan mapping step was removed from the plan feedback and the feedback works on the unmapped $\pi$, plan feedback was removed altogether and where no tools except those for \ac{CRU} are available. All variants were tested on the NL-pddlgym test set with DeepSeek v4 Flash.}
    \label{tab:ablation_comparison}
\end{table}

Furthermore, we also created a variant of the PDDLCoder approach where the same tools and agent are available, but the pipeline is orchestrated in a step-by-step sequence.
Specifically, the pipeline starts by first generating the domain, fixing eventual syntax errors in a loop of up to 5 iterations and goes on to generating the problem with a similar syntax feedback loop.
Once a set of syntactically valid \ac{PDDL} files has been generated, the system tries to generate a plan and starts another feedback loop, where the model is fed the specific issues with instructions to fix the domain and problem files, in cases where the current pair of \ac{PDDL} files is unsolvable.
Once a plan has been generated, it is mapped by the MapAgent and fed into the PlanAgent which generates appropriate feedback.
This feedback is then fed back to the start of the pipeline together with the already created \ac{PDDL} files with instructions to fix any possible issues.
This feedback loop repeats for 5 iterations at which point the pipeline completes, resulting in a minimum of 10 and a maximum of 102 total model calls depending on the number of executed feedback loops.
Although many alternative rigid pipelines could be designed and the interaction budgets are not strictly matched, the result provides preliminary evidence that the adaptive tool selection of PDDLCoder contributes beyond available tools alone.
Specifically, the rigid pipeline variant is outperformed by our proposed pipeline implementation in all tested domains.
While the rigid implementation requires significantly fewer input tokens, as the previous interaction history is not fed into the model on each call, the agentic variant only requires $57.3\%$ of the average number of output tokens consumed by the rigid variant, as the model can apply more focused edits instead of having to regenerate the entire \ac{PDDL} file for each change.

\section{Conclusion}

In this work, we introduce PDDLCoder, an autonomous, agentic LLM-Formalizer comparable to modern coding agents.
By iteratively creating, analyzing, and refining planning specifications, PDDLCoder outperforms our adaptations of both existing LLM-Formalizers and direct LLM-Planners in generating executable, applicable plans on the NL-pddlgym test set.
To support rigorous evaluation in this domain, we also present NL-pddlgym, a benchmark of 711 abstract planning problems that verifies the applicability of plans generated for abstract domain and problem descriptions in executable gym environments.
While PDDLCoder demonstrates strong zero-shot capabilities, the PlanAgent currently relies on sufficient reasoning capacity of \acp{LLM} to correctly identify and resolve logical inconsistencies, while the overall pipeline quality depends on the LLM's general ability to generate syntactically valid and solvable \ac{PDDL}.
To address this, future work will explore automated prompt optimization utilizing the training and validation splits of NL-pddlgym.
Ultimately, we believe this agentic framework and execution-based benchmark establish a strong, reproducible foundation for the next generation of LLM-assisted symbolic planning.

\section{Acknowledgments}
This study was funded by the Deutsche Forschungsgemeinschaft (DFG, German Research Foundation) under the National Research Data Infrastructure – NFDI 2/2 ‘NFDI4Cat – NFDI für Wissenschaften mit Bezug zur Katalyse’ – 441926934.

\bibliography{arxiv}
\clearpage
\appendix

\section{Experimental Settings}

All experimental results were gathered on two systems:
\begin{enumerate}
    \item Laptop running Ubuntu 24.04.4 with an Intel Core I7-12800H and 64 GB of RAM.
    \item GPU server running Ubuntu 24.04.4 with an Intel Xeon Platinum 8480C CPU, 100 GB of RAM and 4 NVIDIA H100 GPUs.
\end{enumerate}

System 1 was used for gathering all results using DeepSeek v4 Flash and system 2 was used to gather the results using the other models.
While the DeepSeek model was queried through OpenRouter \footnote{OpenRouter: available under \url{https://openrouter.ai/}} its temperature was set to $1.0$ and Top P of 1.0 according to the original authors specifications.
All other models were hosted locally using vLLM with the sampling parameters set according to the original model authors' specifications, which are listed in Table \ref{tab:sample-settings}.

\begin{table}[h]
    \centering
    \begin{tabular}{cccc}
        \toprule
        \textbf{Model} & \textbf{Temperature} & \textbf{Top K} & \textbf{Top P} \\
        \midrule
        Gemma 4 31b & 1.0 & 64 & 0.95 \\
        gpt-oss-120b & 1.0 & 0 & 1.0 \\
        Qwen 3.6 35b A3b & 1.0 & 20 & 0.95 \\
        GLM 4.7 Flash & 1.0 & 0 & 1.0 \\
        Llama 4 Scout 17b 16e & 0.6 & 0 & 0.9 \\
        \bottomrule
    \end{tabular}
    \caption{Sampling parameters for all locally hosted models from the experiments.}
    \label{tab:sample-settings}
\end{table}

\section{Supplemental Results}

Note that while performance levels (L0–L3) are reported comprehensively for all models, token consumption statistics for ISR-LLM are omitted in the supplemental tables due to a logging limitation during the local vLLM execution. 
This does not impact the primary applicability and success rate metrics.

\begin{table}[htbp]
    \centering
    \begin{tabular}{l|cccc|cccc|cccc}
        \toprule        
        & \multicolumn{4}{c|}{\textbf{Outcome Counts}} & \multicolumn{4}{c|}{\textbf{Success Rates (\%)}} & \multicolumn{4}{c}{\textbf{Token Statistics}} \\
        \textbf{Approach} & \textbf{L0} & \textbf{L1} & \textbf{L2} & \textbf{L3} & \textbf{El} & \textbf{Ha} & \textbf{RP} & \textbf{Sa} & \textbf{N-In} & \textbf{S-In} & \textbf{N-Out} & \textbf{S-Out} \\
        \midrule
        PDDLCoder & 8 & 2 & 31 & \textbf{65} & 20 & \textbf{85} & 63.3 & 69.4 & 411577.6 & 404900.1 & 15453.8 & 12257.4 \\
        \midrule
        VML\_PDDL & 36 & 0 & 20 & 50 & 60 & 65 & 6.7 & 63.9 & 55211.3 & 34349.9 & 42909.2 & 19945.6 \\
        NL2Plan & 6 & 2 & 37 & 61 & 70 & 75 & 10 & \textbf{80.6} & 128786.3 & 47703.0 & 13668.6 & 6832.4 \\
        \midrule
        ISR-LLM & -- & -- & 58 & 48 & \textbf{85} & 25 & 16.7 & 58.3 & -- & -- & -- & -- \\
        COT & -- & -- & 85 & 21 & 0 & 0 & \textbf{70} & 0 & \textbf{3946.9} & \textbf{4380.1} & \textbf{3336.4} & \textbf{4808.6} \\
        \bottomrule
    \end{tabular}
    \caption{Performance of Gemma 4 31b across PDDLCoder and our adaptations of previous LLM-Formalizers (VML\_PDDL and NL2Plan) and LLM-Planners (ISR-LLM and COT) on the NL-pddlgym test set.}
\end{table}

\begin{table}[htbp]
    \centering
    \begin{tabular}{l|cccc|cccc|cccc}
        \toprule        
        & \multicolumn{4}{c|}{\textbf{Outcome Counts}} & \multicolumn{4}{c|}{\textbf{Success Rates (\%)}} & \multicolumn{4}{c}{\textbf{Token Statistics}} \\
        \textbf{Approach} & \textbf{L0} & \textbf{L1} & \textbf{L2} & \textbf{L3} & \textbf{El} & \textbf{Ha} & \textbf{RP} & \textbf{Sa} & \textbf{N-In} & \textbf{S-In} & \textbf{N-Out} & \textbf{S-Out} \\
        \midrule
        PDDLCoder & 26 & 3 & 21 & \textbf{56} & \textbf{80} & \textbf{55} & 46.7 & \textbf{41.7} & 307318.1 & 424007.3 & 27589.4 & 29793.0 \\
        \midrule
        VML\_PDDL & 73 & 0 & 8 & 25 & 35 & 20 & 26.7 & 16.7 & 59692.1 & 31419.5 & 66750.6 & 19773.4 \\
        NL2Plan & 98 & 2 & 1 & 5 & 20 & 0 & 0 & 2.8 & 265195.9 & 203457.1 & 46588.9 & 33257.3 \\
        \midrule
        ISR-LLM & -- & -- & 60 & 46 & 70 & 20 & \textbf{66.7} & 22.2 & -- & -- & -- & -- \\
        COT & -- & -- & 87 & 19 & 25 & 0 & 46.7 & 0 & \textbf{3663.8} & \textbf{3944.0} & \textbf{4920.7} & \textbf{5009.0} \\
        \bottomrule
    \end{tabular}
    \caption{Performance of gpt-oss-120b across PDDLCoder and our adaptations of previous LLM-Formalizers (VML\_PDDL and NL2Plan) and LLM-Planners (ISR-LLM and COT) on the NL-pddlgym test set}
\end{table}

\begin{table}[htbp]
    \centering
    \begin{tabular}{l|cccc|cccc|cccc}
        \toprule        
        & \multicolumn{4}{c|}{\textbf{Outcome Counts}} & \multicolumn{4}{c|}{\textbf{Success Rates (\%)}} & \multicolumn{4}{c}{\textbf{Token Statistics}} \\
        \textbf{Approach} & \textbf{L0} & \textbf{L1} & \textbf{L2} & \textbf{L3} & \textbf{El} & \textbf{Ha} & \textbf{RP} & \textbf{Sa} & \textbf{N-In} & \textbf{S-In} & \textbf{N-Out} & \textbf{S-Out} \\
        \midrule
        PDDLCoder & 45 & 1 & 49 & \textbf{11} & \textbf{30} & \textbf{5} & 6.7 & \textbf{5.6} & 129916.1 & 251746.6 & 37574.6 & 40290.6 \\
        \midrule
        VML\_PDDL & 106 & 0 & 0 & 0 & 0 & 0 & 0 & 0 & 118813.5 & 43678.3 & 117951.1 & 11979.3 \\
        NL2Plan & 104 & 1 & 1 & 0 & 0 & 0 & 0 & 0 & 99211.5 & 367813.5 & 38490.0 & 139312.0 \\
        \midrule
        ISR-LLM & -- & -- & -- & -- & -- & -- & -- & -- & -- & -- & -- & -- \\
        COT & -- & -- & 101 & 5 & 5 & 0 & \textbf{13.3} & 0 & \textbf{4113.5} & \textbf{5231.2} & \textbf{14464.9} & \textbf{8999.2} \\
        \bottomrule
    \end{tabular}
    \caption{Performance of Qwen 3.6 35b A3b across PDDLCoder and our adaptations of previous LLM-Formalizers (VML\_PDDL and NL2Plan) and LLM-Planners (ISR-LLM and COT) on the NL-pddlgym test set. Note, that results with ISR-LLM could not be generated as the prompting approach used by ISR-LLM, which includes intermediate system prompts, is not supported by this model.}
\end{table}

\begin{table}[htbp]
    \centering
    \begin{tabular}{l|cccc|cccc|cccc}
        \toprule        
        & \multicolumn{4}{c|}{\textbf{Outcome Counts}} & \multicolumn{4}{c|}{\textbf{Success Rates (\%)}} & \multicolumn{4}{c}{\textbf{Token Statistics}} \\
        \textbf{Approach} & \textbf{L0} & \textbf{L1} & \textbf{L2} & \textbf{L3} & \textbf{El} & \textbf{Ha} & \textbf{RP} & \textbf{Sa} & \textbf{N-In} & \textbf{S-In} & \textbf{N-Out} & \textbf{S-Out} \\
        \midrule
        PDDLCoder & 66 & 17 & 22 & \textbf{1} & \textbf{5} & 0 & 0 & 0 & 762909.5 & 721951.2 & 35162.3 & 28381.5 \\
        \midrule
        VML\_PDDL & 105 & 0 & 1 & 0 & 0 & 0 & 0 & 0 & 96135.3 & 40211.3 & 107913.0 & 12473.0 \\
        NL2Plan & 106 & 0 & 0 & 0 & 0 & 0 & 0 & 0 & 217176.2 & 430568.5 & 88899.5 & 168316.3 \\
        \midrule
        ISR-LLM & -- & -- & 106 & 0 & 0 & 0 & 0 & 0 & -- & -- & -- & -- \\
        COT & -- & -- & 105 & \textbf{1} & 0 & 0 & \textbf{3.3} & 0 & \textbf{3654.9} & \textbf{4809.9} & \textbf{8620.3} & \textbf{9011.0} \\
        \bottomrule
    \end{tabular}
    \caption{Performance of GLM 4.7 Flash across PDDLCoder and our adaptations of previous LLM-Formalizers (VML\_PDDL and NL2Plan) and LLM-Planners (ISR-LLM and COT) on the NL-pddlgym test set.}
\end{table}

\begin{table}[htbp]
    \centering
    \begin{tabular}{l|cccc|cccc|cccc}
        \toprule        
        & \multicolumn{4}{c|}{\textbf{Outcome Counts}} & \multicolumn{4}{c|}{\textbf{Success Rates (\%)}} & \multicolumn{4}{c}{\textbf{Token Statistics}} \\
        \textbf{Approach} & \textbf{L0} & \textbf{L1} & \textbf{L2} & \textbf{L3} & \textbf{El} & \textbf{Ha} & \textbf{RP} & \textbf{Sa} & \textbf{N-In} & \textbf{S-In} & \textbf{N-Out} & \textbf{S-Out} \\
        \midrule
        PDDLCoder & 99 & 3 & 4 & 0 & 0 & 0 & 0 & 0 & 521045.1 & 154822.6 & 13961.4 & 6312.5 \\
        \midrule
        VML\_PDDL & 106 & 0 & 0 & 0 & 0 & 0 & 0 & 0 & 48656.2 & 31920.6 & 32228.2 & 13921.2 \\
        NL2Plan & 76 & 11 & 14 & \textbf{5} & 10 & \textbf{5} & \textbf{6.7} & 0 & 164695.2 & 89894.1 & 15800.3 & 9109.9 \\
        \midrule
        ISR-LLM & -- & -- & 101 & \textbf{5} & 10 & \textbf{5} & 3.3 & \textbf{2.8} & -- & -- & -- & -- \\
        COT & -- & -- & 101 & \textbf{5} & \textbf{15} & 0 & \textbf{6.7} & 0 & \textbf{2814.9} & \textbf{2199.8} & \textbf{893.2} & \textbf{517.6} \\
        \bottomrule
    \end{tabular}
    \caption{Performance of Llama 4 Scout 17b 16e across PDDLCoder and our adaptations of previous LLM-Formalizers (VML\_PDDL and NL2Plan) and LLM-Planners (ISR-LLM and COT) on the NL-pddlgym test set.}
\end{table}

\FloatBarrier
\section{Example of a Successful PDDLCoder Run}

The following presents a successful PDDLCoder execution trace for a problem in the Ring and Peg domain.
The task starts out with the following $D_{NL}$ and $P_{NL}$

\begin{SmallInstructionbox}{$D_{NL}$ and $P_{NL}$}
### Domain:
## Domain description

A robot arm is tasked to sort a set of colored rings on a set of pegs of the same colors. The arm can be used to pick up a ring from a peg, move a ring to another peg and place the ring on the other peg. Rings are already placed on pegs and can only be placed on pegs. To pick up a ring at a peg, the robot first has to move to the pegs position.

The actions available to the robot are:
- move
    - Moves the robot to a specified position
- pick
    - Closes the robot arms gripper at the current position
- place
    - Opens the robot arms gripper at the current position


### Problem:
## Problem description

There are 5 colored pegs with the following colors: red, green, blue, pink, yellow. The pegs are all named in the format <color>_peg. Additionally there are 4 colored rings: red, green, yellow and pink. Similarly to the pegs, the rings are named <color>_ring.

The robot arm starts out in a default position, while the rings start in the following positions:
- red_ring - red_peg
- green_ring - yellow_peg
- yellow_ring - blue_peg
- pink_ring - green_peg

The goal is to transfer all rings to the pegs of their color. 
\end{SmallInstructionbox}

PDDLCoder then initially creates the following files, which are verified to be syntactically correct after creation.

\begin{SmallInstructionbox}{$D$}
(define (domain ring-sorting)
    (:requirements :strips :typing :negative-preconditions :equality)
    (:types
        location
        peg - location
        ring
    )  
    (:predicates
        (robot_at ?l - location)
        (hand_empty)
        (holding ?r - ring)
        (on_peg ?r - ring ?p - peg)
    )
    (:action move
        :parameters (?from ?to - location)
        :precondition (and (robot_at ?from) (not (= ?from ?to)))
        :effect (and (robot_at ?to) (not (robot_at ?from)))
    )
    (:action pick
        :parameters (?r - ring ?p - peg)
        :precondition (and (robot_at ?p) (hand_empty) (on_peg ?r ?p))
        :effect (and (holding ?r) (not (hand_empty)) (not (on_peg ?r ?p)))
    )
    (:action place
        :parameters (?r - ring ?p - peg)
        :precondition (and (robot_at ?p) (holding ?r))
        :effect (and (on_peg ?r ?p) (hand_empty) (not (holding ?r)))
    )
)
\end{SmallInstructionbox}

\begin{SmallInstructionbox}{$P$}
(define (problem sort-rings)
    (:domain ring-sorting)
    (:objects
        default_pos - location
        red_peg green_peg blue_peg pink_peg yellow_peg - peg
        red_ring green_ring yellow_ring pink_ring - ring
    )
    (:init
        (robot_at default_pos)
        (hand_empty)
        (on_peg red_ring red_peg)
        (on_peg green_ring yellow_peg)
        (on_peg yellow_ring blue_peg)
        (on_peg pink_ring green_peg)
    )
    (:goal
        (and
            (on_peg red_ring red_peg)
            (on_peg green_ring green_peg)
            (on_peg yellow_ring yellow_peg)
            (on_peg pink_ring pink_peg)
        )
    )
)
\end{SmallInstructionbox}

The LLM then goes on to call the tool for plan generation, which generates the following plan.

\begin{SmallInstructionbox}{$\pi$}
(move default_pos blue_peg)
(pick yellow_ring blue_peg)
(move blue_peg green_peg)
(move green_peg yellow_peg)
(place yellow_ring yellow_peg)
(pick green_ring yellow_peg)
(move yellow_peg green_peg)
(place green_ring green_peg)
(pick pink_ring green_peg)
(move green_peg pink_peg)
(place pink_ring pink_peg)
; cost = 11 (unit cost)
\end{SmallInstructionbox}

To verify that $\pi$ is correctly aligned with $D_{NL}$ and $P_{NL}$, the LLM then goes on to call the PlanAgent, which returns the following feedback.

\begin{SmallInstructionbox}{Plan Feedback}
The plan is logically flawed and physically infeasible. After moving to yellow_peg and placing the yellow_ring, the next action (pick) at the same peg would pick the topmost ring, which is the just-placed yellow_ring, not the green_ring that needs to be moved. This causes the subsequent moves to place the yellow_ring incorrectly on green_peg and then pink_peg. The plan cannot achieve the goal because it lacks a mechanism to select a specific ring from a peg with multiple rings. Additionally, the detour to green_peg before yellow_peg is unnecessary but not the main issue.
\end{SmallInstructionbox}

In response, the LLM then generates the following output and goes on to apply the fix to $D$ and $P$, which are shown in their updated form below.
While the noted fix is not the only valid approach to the problem, as it avoids the infeasibility of picking a lower ring from a stack by simply disallowing ring stacking, this approach still results in an applicable plan for the given problem.

\begin{SmallInstructionbox}{Thought}
The plan feedback indicates that the current domain allows picking any ring from a peg regardless of stack order, which is physically infeasible. To fix this, we should enforce that each peg can hold at most one ring. This avoids stacking issues and matches the problem well, as there are 5 pegs and only 4 rings, leaving one peg empty. We need to add a predicate `(peg_empty ?p - peg)` and modify the `place` action to require the destination peg to be empty, and the `pick` action to make the peg empty after picking. Also update the problem file with initial and goal `peg_empty` states. We'll edit the domain and problem files accordingly. 
\end{SmallInstructionbox}

\begin{SmallInstructionbox}{Updated $D$}
(define (domain ring-sorting)
    (:requirements :strips :typing :negative-preconditions :equality)
    (:types
        location
        peg - location
        ring
    )
    (:predicates
        (robot_at ?l - location)
        (hand_empty)
        (holding ?r - ring)
        (on_peg ?r - ring ?p - peg)
        (peg_empty ?p - peg)
    )
    (:action move
        :parameters (?from ?to - location)
        :precondition (and (robot_at ?from) (not (= ?from ?to)))
        :effect (and (robot_at ?to) (not (robot_at ?from)))
    )
    (:action pick
        :parameters (?r - ring ?p - peg)
        :precondition (and (robot_at ?p) (hand_empty) (on_peg ?r ?p))
        :effect (and (holding ?r) (not (hand_empty)) (not (on_peg ?r ?p)) (peg_empty ?p))
    )
    (:action place
        :parameters (?r - ring ?p - peg)
        :precondition (and (robot_at ?p) (holding ?r) (peg_empty ?p))
        :effect (and (on_peg ?r ?p) (hand_empty) (not (holding ?r)) (not (peg_empty ?p)))
    )
)
\end{SmallInstructionbox}

\begin{SmallInstructionbox}{Updated $P$}
(define (problem sort-rings)
    (:domain ring-sorting)
    (:objects
        default_pos - location
        red_peg green_peg blue_peg pink_peg yellow_peg - peg
        red_ring green_ring yellow_ring pink_ring - ring
    )
    (:init
        (robot_at default_pos)
        (hand_empty)
        (peg_empty pink_peg)
        (on_peg red_ring red_peg)
        (on_peg green_ring yellow_peg)
        (on_peg yellow_ring blue_peg)
        (on_peg pink_ring green_peg)
    )
    (:goal
        (and
            (on_peg red_ring red_peg)
            (on_peg green_ring green_peg)
            (on_peg yellow_ring yellow_peg)
            (on_peg pink_ring pink_peg)
        )
      )
)
\end{SmallInstructionbox}

As both files are syntactically correct and solvable, they yield the following $\pi$ generated by the planning tool.

\begin{SmallInstructionbox}{Updated $\pi$}
(move default_pos blue_peg)
(move blue_peg green_peg)
(pick pink_ring green_peg)
(move green_peg blue_peg)
(move blue_peg pink_peg)
(place pink_ring pink_peg)
(move pink_peg blue_peg)
(move blue_peg yellow_peg)
(pick green_ring yellow_peg)
(move yellow_peg green_peg)
(place green_ring green_peg)
(move green_peg blue_peg)
(pick yellow_ring blue_peg)
(move blue_peg yellow_peg)
(place yellow_ring yellow_peg)
; cost = 15 (unit cost)
\end{SmallInstructionbox}

To verify that the generated $\pi$ is properly aligned with the NL descriptions, the LLM then calls the PlanAgent once more and finishes execution after 12 iterations, as the PlanAgent states that the generated $\pi$ is properly aligned.

\begin{SmallInstructionbox}{Plan Feedback on Updated $\pi$}
The plan is logically and physically feasible. It correctly moves the pink ring from green peg to pink peg, the green ring from yellow peg to green peg, and the yellow ring from blue peg to yellow peg. All picks and places occur at pegs that contain the intended ring or are empty, respectively. The sequence of moves allows the arm to reach the necessary positions while holding rings appropriately. No logical or physical constraints are violated. 
\end{SmallInstructionbox}

After the execution has succeeded, $D$ and $P$ are externally checked for syntactic validity, $\pi$ is regenerated, and the MapAgent finally generates the following applicable $\pi_m$.

\begin{SmallInstructionbox}{$\pi_m$}
(move blue_peg)
(move green_peg)
(pick)
(move blue_peg)
(move pink_peg)
(place)
(move blue_peg)
(move yellow_peg)
(pick)
(move green_peg)
(place)
(move blue_peg)
(pick)
(move yellow_peg)
(place)
\end{SmallInstructionbox}

\section{PDDLCoder Prompts}

\begin{SmallInstructionbox}{PDDLCoder Prompt}
## Task Context

You are an expert at generating PDDL domain and problem files. You will be asked to generate PDDL files for a task given in natural language.
Make sure to adhere to the naming of objects, entities and actions given in the prompt. There are no other objects in the domain and the robot is only able to carry out the given actions. No other actions than the ones explicitly listed for a task are allowed in the domain and plan.
Additionally, make sure your output is complete and can be used by a solver to generate a plan once both domain and problem files have been generated. This means including a full definition of the actions, predicates and other PDDL attributes you deem necessary to complete the task.
After creating the files, verify that they do not contain syntax mistakes. If they do contain mistakes, edit the files to fix them. Additionally, ensure that the task defined in the PDDL files is actually solvable using the Fast Downward planning system. Before a plan can be generated, ensure that the translation layer of the Fast Downward planning system runs without reporting any further issues. Once a plan has been generated ensure it is actually physically and logically feasible for the given task by getting feedback from the get_plan_feedback function. **Always** use and incorporate the feedback from the get_plan_feedback function before completing your task. The generated plan needs to satisfy the described physical domain, not only the PDDL files you generated.
All PDDL files have to adhere to the PDDL 1.2 standard. Do not use constants or derived predicates. You may use the following requirements: strips, disjunctive-preconditions, conditional-effects, negative-preconditions
\end{SmallInstructionbox}

\begin{SmallInstructionbox}{MapAgent Prompt}
# Role

You are an expert of mapping a plan of PDDL actions to a plan with a different action schema and object naming.
You will be given the original plan of PDDL actions, the corresponding PDDL domain with the definitions of the plans actions, a definition of the actions that the plan sequence should be mapped to and lastly a list of the object names available in the environment.
Make sure to fit the actions parameters and object names to the semantically closest parameters of the new action schema. If the new action schema does not contain any parameters, you do not need to map any parameters or object names.
If the plan contains actions not given in the schema, leave those actions in the plan as is.
\end{SmallInstructionbox}

Note, that the same MapAgent prompt is being used in our adaptations of VML\_PDDL and NL2Plan when mapping the generated plans to their simulation-ready form.

\begin{SmallInstructionbox}{PlanAgent Prompt}
# Role

You are an expert at providing feedback for the feasibility and applicability of a given plan. You will be provided with a description of the plan that should be solved, the plan and the schema of the involved actions.

**Your focuses are**:
- Does the sequence of actions logically lead to the desired outcomes?
- Is the sequence of actions physically and logically feasible?

The plan will need to be carried out in the real world and should therefore respect real logical and physical constraints.

Do not make suggestions apart from the focuses noted above.

**Important**:
- The plan does not need to be optimal as long as it solves the given task. Do not provide feedback on redundant moves.
- Ensure that the plan only uses the actions given in the task description.
- Be specific in your critique and provide suggestions to fix logical/physical flaws.
- Do not try to provide a plan yourself.
- Provide actional feedback on why a plan is feasible/unfeasible.
- The cost annotations at the end of the file are an artifact and have no relevance for your feedback.
- BE CONCISE.
\end{SmallInstructionbox}

\section{Rigid PDDLCoder Prompts}

\begin{SmallInstructionbox}{Rigid PDDLCoder Prompt}
## Task Context

You are an expert at generating PDDL domain and problem files. You will be asked to generated PDDL files for a task given in natural language. You will first be asked to generate the domain file for the given task.
Make sure to adhere to the naming of objects and actions given in the prompt.
Additionally, make sure your output is complete and can be used by a solver to generate a plan once both domain and problem files have been generated. This means including a full definition of the actions, predicates and other PDDL attributes you deem necessary to complete the task.
All PDDL files have to adhere to the PDDL 1.2 standard. If possible, refrain from defining :constants in the domain. It should be possible to define multiple problems with different objects for the same domain.
Answer **only** with PDDL as output. The result will directly be used in a PDDL planning system and should therefore be syntactically and semantically sound.
\end{SmallInstructionbox}

\begin{SmallInstructionbox}{Syntax Error Prompt}
# Task Context

You are an expert at fixing PDDL files.
You will be given PDDL domain or problem files, which contain syntax errors.
In addition to the PDDL files, you will also be presented with the list of errors in the given file.
The errors will present the original line from the PDDL file and the corresponding error message.
You will need to fix the syntax mistakes and return the fixed file.
Answer **only** with PDDL as output. The result will directly be used in a PDDL planning system and should therefore be syntactically and semantically sound.
All PDDL files have to adhere to the PDDL 1.2 standard.
\end{SmallInstructionbox}

\begin{SmallInstructionbox}{Translation Error Prompt}
## Task Context

You are an expert at fixing syntactically invalid PDDL domain and problem files. You will be asked to fix and improve a given pair of PDDL domain and problem files that currently contain syntax mistakes. You will be given information on which of thet two you should fix, as only one of the two files can be changed at a time.
All PDDL files have to adhere to the PDDL 1.2 standard.
Answer **only** with PDDL as output. The result will directly be used in a PDDL planning system and should therefore be syntactically and semantically sound. 
\end{SmallInstructionbox}

\begin{SmallInstructionbox}{Planning Error Prompt}
## Task Context

You are an expert at fixing unsolvable PDDL domain and problem files. You will be asked to fix and improve a given pair of PDDL domain and problem files that are currently unsolvable. You will be prompted with which of the two you should be improving, as only one of the two files should be changed at a time.
Answer **only** with PDDL as output. The result will directly be used in a PDDL planning system and should therefore be syntactically and semantically sound.
All PDDL files have to adhere to the PDDL 1.2 standard.
\end{SmallInstructionbox}

\section{COT Prompts}

\begin{SmallInstructionbox}{COT Prompt}
## Task Context

You are an expert at generating plans for solving a given problem.
Make sure to adhere to the naming of objects, entities and actions given in the prompt. There are no other objects in the domain and the robot is only able to carry out the given actions. No other actions than the ones explicitly listed for a task are allowed in the domain and plan.
Additionally, make sure your output is complete and actually solves the given problem. Note, that for the plan to be correct, each of the actions and their parameters in the plan need to be written to a new line in the plan, surrounded by parantheses. So for an action 'action' with parameters 'p1', 'p2' the resulting step in the plan would be '(action p1 p2)'.    
\end{SmallInstructionbox}

\begin{SmallInstructionbox}{MapAgent Prompt}
# Role

You are an expert of mapping a plan of PDDL actions to a plan with a different action schema and object naming.
You will be given the original plan of PDDL actions, a definition of the actions that the plan sequence should be mapped to and lastly a list of the object names available in the environment.
Make sure to fit the actions parameters and object names to the semantically closest parameters of the new action schema. If the new action schema does not contain any parameters, you do not need to map any parameters or object names.
If the plan contains actions not given in the schema, leave those actions in the plan as is.
\end{SmallInstructionbox}

Note, that the MapAgent prompt that is being used for the COT implementation is the same for the final plan mapping that is being used in our adaptation of ISR-LLM. 

\section{Adapted Related Work Prompts}

To run VML\_PDDL with the remotely hosted DeepSeek model, we adapated the original approach to run with LLM-based grading instead of the original log-likelihood based grading system.
Following is the prompt used for this grading step.
All other prompts that are used in the related work are the same as in the original implementations.

\begin{SmallInstructionbox}{LLM-Based Grading of Generated Domains for VML\_PDDL}
Score the following PDDL domain on how well it captures the semantics of the NL description (0-100). Return only the number.

NL: {nl_description}

PDDL Domain:
{domain}

Score:
\end{SmallInstructionbox}

\begin{SmallInstructionbox}{LLM-Based Grading of Generated Problems for VML\_PDDL}
Score the following PDDL problem on how well it captures the semantics of the NL description and is consistent with the given domain (0-100). Return only the number.

NL: {nl_description}

Domain:
{domain}

Problem:
{problem}

Score:    
\end{SmallInstructionbox}

\section{NL-pddlgym Test Set Domain and Problem Descriptions}

\begin{SmallInstructionbox}{Elevator Domain Description}
## Domain Description

A robot is tasked with driving an elevator to pick up people and drive them to their desired floor.

The available actions are:
- down
    - Moves the elevator down 1 floor
- up
    - Moves the elevator up 1 floor
- board
    - Lets the passenger board the elevator
- depart
    - Lets the passenger depart the elevator
\end{SmallInstructionbox}

\begin{SmallInstructionbox}{Exemplary Elevator Problem Description}
## Problem Description

There are six floors (f0 - f5) and three passengers (p0 - p2). The floors are arranged in the following order (bottom to top): f0,f1,f2,f3,f4,f5. The lift is currently at floor f0. The passenger p0 is at floor f1, p1 is at f3 and p2 is at f5. The goal is to transfer the passenger p0 to floor f4, p1 to f1 and p2 to f1.
\end{SmallInstructionbox}

\begin{SmallInstructionbox}{Hanoi Domain Description}
## Domain description

A robot is tasked with stacking a set of rings of different sizes on a set of pegs such that the rings are stacked on one peg with the largest ring of the bottom with each subsequent ring being the next smaller available ring.

The actions available to the robot are:
- move
    - Moves a ring from one peg to the other.
\end{SmallInstructionbox}

\begin{SmallInstructionbox}{Exemplary Hanoi Problem Description}
## Problem description

There are 3 pegs (peg1 - peg3) and 4 rings (d1 - d4). The rings have the following order according to their size (largest to smallest): d4,d3,d2,d1. peg2 and peg3 have no rings stacked on them. peg1 has rings stacked on it in the following order (bottom to top): d4,d3,d2,d1. The rings should be stacked from largest to smallest on peg peg3.
\end{SmallInstructionbox}

\begin{SmallInstructionbox}{Satellite Domain Description}
## Domain description

A robot is tasked with controlling a set of satellites to capture images. Each satellite can have different instruments with different capabilities on board.

The actions available to the robot are:
- turn_to
    - Points a satellite in a specific direction.
- switch_on
    - Turns on a specific instrument on the satellite. Once an instrument is switched on on a satellite, all power is consumed by this instrument and the power only becomes available again once the instrument is turned off.
- switch_off
    - Turns off a specific instrument on the satellite.
- calibrate
    - Calibrates an instrument for a specific target direction.
- take_image
    - Takes an image in a specific mode of the current direction of the satellite using one a calibrated instrument.
\end{SmallInstructionbox}

\begin{SmallInstructionbox}{Exemplary Satellite Problem Description}
## Problem description

There are two satellites (satellite0, satellite1), four instruments (instrument0 - instrument3), three different image modes (image1, spectograph2, infrared0) and eight directions (Star0 - Star4, Phenomenon5 - Phenomenon7). The satellites satellite0 and satellite1 have power available. The satellite satellite0 carries the instruments instrument0, instrument1 and instrument2 and satellite1 carries instrument3. The instrument instrument0 support the following modes: spectograph2, infrared0. instrument1 is supports image1. instrument2 supports infrared0 and image1. instrument3 supports spectograph2, infrared0 and image1. instrument0 is currently callibrated for the Star1, instrument1 for Star2, instrument2 for Star0 and instrument3 for Star0. The satellite satellite0 is currrently pointing in the direction of Star4 and satellite1 is pointing to Star0. The goal is to capture an image of Star3 with infrared0, of Star4 with spectograph2, of Phenomenon5 with spectograph2, of Phenomenon7 with spectograph2 and for satellite0 to point to Phenomenon5.
\end{SmallInstructionbox}

\end{document}